\documentclass[letterpaper, 10 pt, conference]{ieeeconf}  % Comment this line out if you need a4paper

\IEEEoverridecommandlockouts % This command is only needed if 
\usepackage{amsmath}
\usepackage{amssymb}
\usepackage{subfiles}
\usepackage{bm}
\usepackage{color}
\usepackage{xspace}
\usepackage[bookmarks=true]{hyperref}
\usepackage{enumerate}
\usepackage{graphicx}
\usepackage{caption}
\usepackage{subfig}
\usepackage{multirow}
\usepackage{graphicx}
\usepackage{balance}

\DeclareMathOperator*{\argmax}{arg\,max}
\DeclareMathOperator*{\argmin}{arg\,min}

\newcommand*{\Rthree}{\ensuremath{\mathbb{R}^3}}

\newcommand*{\Rseven}{\ensuremath{\mathbb{R}^7}}

\newcommand*{\xstar}{\ensuremath{\bm x^\ast}}
\newcommand*{\Aone}{\ensuremath{\mathcal{A}_1}}
\newcommand*{\Atwo}{\ensuremath{\mathcal{A}_2}}
\newcommand*{\Ddiff}{\ensuremath{\mathcal{D}}}
\newcommand*{\suppA}{\ensuremath{\sigma_{\mathcal{A}}}}
\newcommand*{\suppAone}{\ensuremath{\sigma_{\mathcal{A}_1}}}
\newcommand*{\suppAtwo}{\ensuremath{\sigma_{\mathcal{A}_2}}}
\newcommand*{\suppDdiff}{\ensuremath{\sigma_{\mathcal{D}}}}

\newcommand{\norm}[1]{\left\lVert#1\right\rVert}

\title{\LARGE \bf
Differentiable Collision Detection: a Randomized Smoothing Approach
}

\author{Louis Montaut, Quentin Le Lidec, Antoine Bambade, Vladimir Petrik, Josef Sivic and Justin Carpentier}

\begin{document}

\maketitle
\thispagestyle{empty}
\pagestyle{empty}

\begin{abstract}
    Collision detection appears as a canonical operation in a large range of robotics applications from robot control to simulation, including motion planning and estimation.
    While the seminal works on the topic date back to the 80s, it is only recently that the question of properly differentiating collision detection has emerged as a central issue, thanks notably to the ongoing and various efforts made by the scientific community around the topic of differentiable physics.
    Yet, very few solutions have been suggested so far, and only with a strong assumption on the nature of the shapes involved.
    In this work, we introduce a generic and efficient approach to compute the derivatives of collision detection for \textit{any} pair of convex shapes, by notably leveraging randomized smoothing techniques which have shown to be particularly adapted to capture the derivatives of non-smooth problems.
    This approach is implemented in the HPP-FCL and Pinocchio ecosystems, and evaluated  on classic datasets and problems of the robotics literature, demonstrating few micro-second timings to compute informative derivatives directly exploitable by many real robotic applications including differentiable simulation.
\end{abstract}
\section{Introduction}
\label{sec:introduction}

Collision detection is a crucial stage for many robotic applications including motion planning, trajectory optimization, or simulation.
It is also used in many other related domains, such as computer graphics or computational geometry, just to name a few.
In particular for simulation, collision detection is a central component of any physical simulator, allowing to assess the collision between two geometries, to retrieve the contact regions if any, or to compute the closest points between the shapes (also known as \textit{witness points}~\cite{ericsonRealTimeCollisionDetection}).
In addition to the evaluation of collisions, many problems in robotics also rely on the derivatives of geometric contact information such as optimal shape design, grasp synthesis, or differentiable simulation.

Over the past few years, differentiable simulation has gained interest within the robotics and machine learning communities thanks to the progress in gradient-based optimization techniques.
Differentiable simulation consists in evaluating the gradient (or sub-gradient in case of a non-smooth contact interaction) of the simulation steps, which can then be exploited by any gradient-based optimization algorithm to efficiently solve various robotic problems involving contact interactions~\cite{de2018end,carpentier2018analytical,toussaint_differentiable_2018,lelidec2021differentiable,geilinger_add_2020,werling_fast_2021}.
However,
most of the existing works on the topic have only considered the differentiation of the physical principles (Coulomb friction constraints, maximum dissipation principles, Signorini conditions, etc.), without systematically considering the contribution of collision detection in the derivatives, for example, by limiting themselves to a specific type of shapes~\cite{toussaint_differentiable_2018,werling_fast_2021}.
%
 
%%%%%%%%%% FIGURE %%%%%%%%%%
%%%%%%%%%%%%%%%%%%%%%%%%%%%%
\begin{figure}[!t]
    \vspace{-0.5cm}
    \centering
    \includegraphics[width=0.98\linewidth]{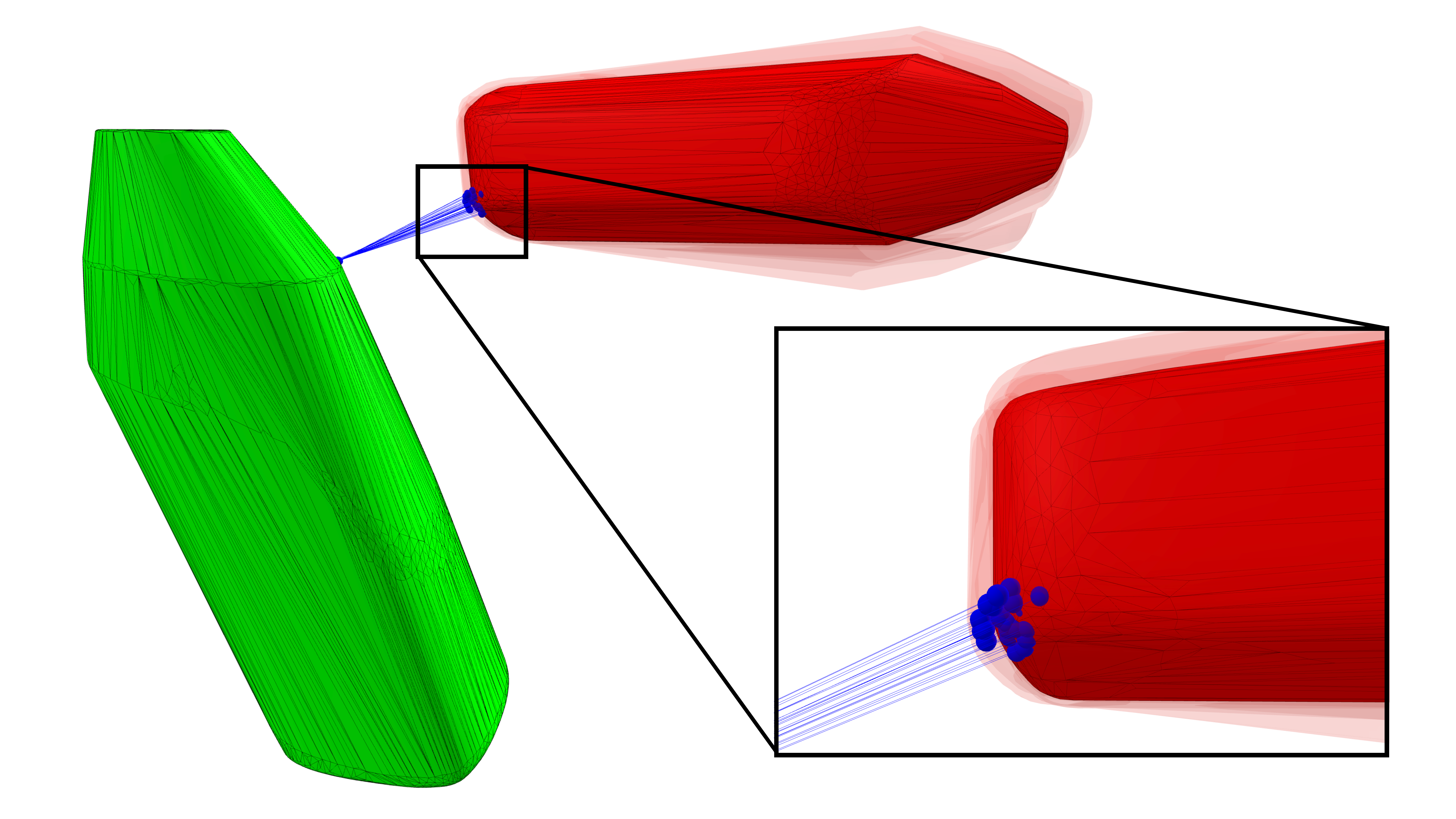}
    \caption{\small\textbf{Illustration of randomized smoothing approximation on two meshes from the YCB dataset. $0^{\text{th}}$-order method, Gaussian distribution,~$M=25$ samples.} 
    The witness points for each perturbed relative pose are drawn on both objects.
    Randomized smoothing allows to recover information regarding the curvature of the underlying objects.
    }
    \label{fig:zero_order_sampling}
    \vspace{-0.5cm}
\end{figure}
%%%%%%%%%%%%%%%%%%%%%%%%%%%%
%%%%%%%%%%%%%%%%%%%%%%%%%%%%
 
To the best of our knowledge, no systematic procedure to compute collision derivatives has been proposed in the literature and it remains a challenging problem. 
In this paper, we propose to address this issue by introducing a generic approach to compute gradient information of collision detection instances in the context of convex shapes (not necessarily strictly convex), without assuming any regularity on the shapes.
This includes not only the standard geometric primitives (ellipsoid, sphere, cone, cube, cylinder, capsules, etc.) but also meshes, which are the standard representation of objects in many applications involving physics simulation and a wide range of robotic applications.
Our key contributions lie in leveraging randomized smoothing techniques~\cite{NEURIPS2020_6bb56208} to estimate the gradients of the collision detection procedure through two proposed estimators:
\begin{itemize}
    \item a $0^\text{th}-$order estimator (see Fig.~\ref{fig:zero_order_sampling}), which does not make any assumption on the nature of the collision algorithms in use,
    \item a $1^\text{st}-$order estimator (see Fig.~\ref{fig:first_order_gumbel_neighbors}), which exploits the optimality conditions of the underlying optimization program inherent to collision detection with convex shapes. Remarkably, this estimator is at the same time more accurate and less expensive to derive than the~$0^{\text{th}}$-order estimator.
\end{itemize}

After stating the problem of collision detection in Sec.~\ref{sec:problem_statement}, we evaluate the efficiency of these two derivative estimators introduced in Sec.~\ref{sec:zero_order_diff} and~\ref{sec:first_order_diff} on standard optimization problems (Sec.~\ref{sec:experiments}).
We will release our code as open-source within the Pinocchio~\cite{carpentier2019pinocchio} framework, so that it can be easily disseminated into other existing robotic software and applications.
\section{Problem statement}
\label{sec:problem_statement}

Collision detection is a very old topic within the robotics community. It consists in determining whether a pair of shapes are in collision or not, and in finding the contact point locations~\cite{ericsonRealTimeCollisionDetection,GJK88,van_den_bergen_proximity_2001,montaut_collision_2022}.
To lower the computational burden, the shapes are often assumed to be convex or decomposed as a collection of convex meshes~\cite{mamouSimpleEfficientApproach2009}.
Such convexity assumptions have led to highly efficient, generic and robust algorithms to solve collision detection problems, most of which belong to the family of the Gilbert-Johnson-Keerthi (GJK) algorithms~\cite{GJK88,van_den_bergen_proximity_2001,mpr2008,montaut_collision_2022}.

From an optimization perspective, the problem of collision detection can be generically formulated as a convex minimization program of the form:
\begin{equation}
    \bm x^*_1, \bm x^*_2 = \argmin_{\bm x_1 \in \Aone,~\bm x_2 \in \Atwo} \norm{\bm x_1 - \bm x_2}_2^2,
    \label{eq:collision_detection_problem}
\end{equation}
where~$\norm{\cdot}_2$ is the Euclidean norm, $\Aone$ and~$\Atwo$ are convex shapes and $\bm x^*_1$ and $\bm x^*_2$ are the so-called witness points.
The GJK algorithm is typically used to solve~\eqref{eq:collision_detection_problem}.
When the objects are in collision, most physics simulations extend~\eqref{eq:collision_detection_problem} to also recover the~\textit{penetration depth}, in order, for instance, to correct the interpenetration between rigid objects, inherent to the discrete integration scheme used inside physical simulators.
The penetration depth corresponds to the length of the smallest translation which must be applied on the relative configuration between ~$\Aone$ and $\Atwo$ in order to separate them.
To compute this quantity, the Expanded Polytope Algorithm (EPA)~\cite{ericsonRealTimeCollisionDetection} is typically used and enables the computation of witness points~$\bm x^*_1$ and~$\bm x^*_2$ in the case of penetration. 
In both penetration and non-collision cases, the witness points computed by GJK and EPA lie on the boundaries of the objects considered.
Finally, in the presence or absence of collision, the signed distance between~$\Aone$ and~$\Atwo$ is defined as:
\begin{equation}
 d(\Aone, \Atwo) = 
 \begin{cases}
      \norm{\bm x^*_1 - \bm x^*_2} \text{ if } \Aone \cap \Atwo = \emptyset, \\
       -\norm{\bm x^*_1 - \bm x^*_2} \text{ otherwise.}
     \end{cases}
 \end{equation}
 
To parameterize the relative position and orientation of~$\Aone$ and~$\Atwo$ in the 3D space, it is convenient to consider their relative configuration,~$T(\bm q) \in SE(3)$ which is parameterized by a vector~$\bm q \in \Rseven$
(3 coordinates for translation and 4 for rotation, encoded by a quaternion).
From a mathematical perspective, the main objective of this paper is to retrieve the partial derivatives of~\eqref{eq:collision_detection_problem} w.r.t~$\bm q$ namely~$\partial \bm x^*_1 / \partial \bm q$ and~$\partial \bm x^*_2 / \partial \bm q$, in the two cases where the two geometries are in collision or not. 
It is worth mentioning at this stage that~$\partial \bm x^*_1 / \partial \bm q$ and~$\partial \bm x^*_2 / \partial \bm q$ are Jacobian matrices of dimension $3 \times 6$, with $\partial \bm q$ being an element of the tangent of $SE(3)$ in $T(\bm q)$~\cite{sola_lie_2021}.

Finite differences and other existing methods are likely to fail at capturing informative gradients in most cases.
Indeed, even if an object is originally smooth, its approximation as a mesh may present some non smoothness properties (typically localised around the vertices of the mesh), which are difficult to handle for classic optimization-based techniques.
More precisely, because they are locally flat or non-smooth, the resulting gradients tend to~\textit{"forget"} the curvature of the original surface they approximate~\cite{werling_fast_2021}, leading to uninformative gradients.
To overcome these issues, we introduce hereafter two approaches to retrieve a precise estimator of the partial derivatives~$\partial \bm x^*_1 / \partial \bm q$ and~$\partial \bm x^*_2 / \partial \bm q$ which are extensively cross-validated against finite differences in Sec.~\ref{sec:experiments}.
\section{$0^{\text{TH}}$-order estimation of collision detection derivatives}
\label{sec:zero_order_diff}
In this section, we introduce a generic approach to compute a $0^\text{th}$-order estimate of the collision detection derivatives by leveraging randomized smoothing techniques which we first review.
\subsection{Background on randomized smoothing}
Originally used in black-box optimization algorithms~\cite{matyas1965random, polyak1987randomopt}, randomized smoothing was recently introduced in the machine learning community~\cite{duchi2012randomized,NEURIPS2020_6bb56208} in order to integrate discrete operations inside neural networks. 

%%%%%%%%%% FIGURE %%%%%%%%%%
%%%%%%%%%%%%%%%%%%%%%%%%%%%%
\begin{figure}[!t]
    \vspace{-0.5cm}
    \centering
    \includegraphics[width=0.9\linewidth]{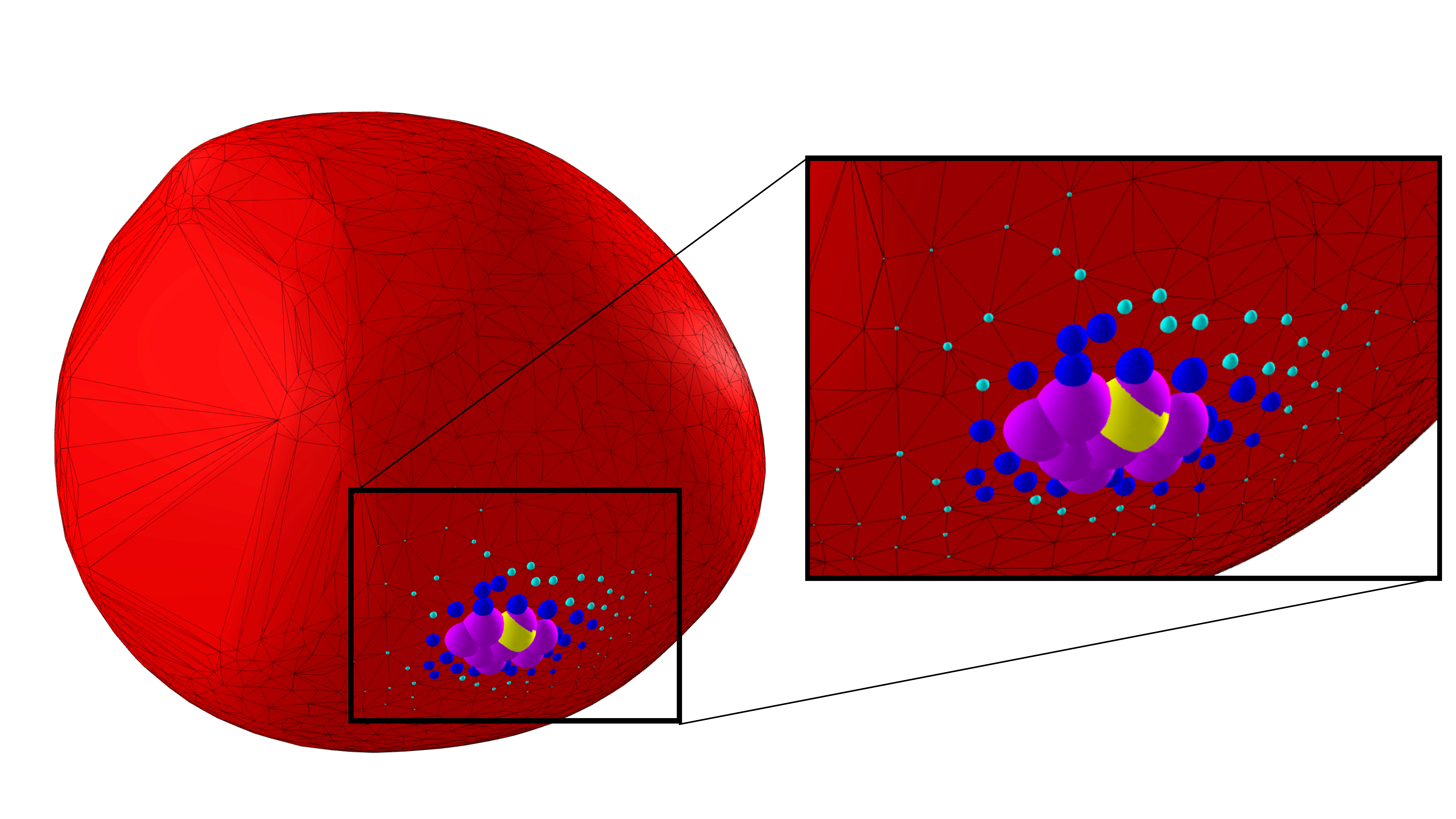}
    \caption{\small\textbf{Soft-max weights of the Gumbel distribution for the~$1^{\text{st}}$-order gradient estimator, mesh from the YCB dataset.
    } 
    Different colors represent neighbors of different levels.
    In yellow: the current-witness point of the object.
    In purple: level $1$ neighbors.
    In blue: level $3$ neighbors.
    In cyan: level $5$ neighbors.
    The size of the neighbors is proportional to the soft-max weight.
    }
    \label{fig:first_order_gumbel_neighbors}
    \vspace{-0.5cm}
\end{figure}
%%%%%%%%%%%%%%%%%%%%%%%%%%%%
%%%%%%%%%%%%%%%%%%%%%%%%%%%%

In more details, any function $g$ can be approximated by convolving it with a probability distribution $\mu$:
\begin{equation}
    g_\epsilon(x) = \mathbb{E}_{Z\sim \mu} \left[ g(x + \epsilon Z) \right],
    \label{eq:rs_smooth_function}
\end{equation}
which corresponds to the randomly smoothed counterpart of $g$ and which can be estimated with a Monte-Carlo estimator as follows:
\begin{equation}
    g_\epsilon(x) \approx \frac{1}{M} \sum_{i=0}^M g(x + \epsilon z^{(i)}),
\end{equation}
where $\{ z^{(1)}, \dots, z^{(M)}\}$ are i.i.d. samples and $M$ is the number of samples.
The~$\epsilon$ parameter controls the level of noise injected in~$g_\epsilon$.
Intuitively, the convolution makes $g_\epsilon$ smoother than its original counterpart $g$ and, thus, yields better conditioned gradients~\cite{NEURIPS2020_6bb56208}. 

Using an integration by part yields a~$0^{\text{th}}$-order estimator of the gradient:
\begin{equation}
     {\nabla^{(0)}_x \, g_\epsilon(x)} =  \frac{1}{M} \sum_{j=0}^M -g(x + \epsilon z^{(j)}) \frac{\nabla \log \mu (z^{(j)})^\top}{\epsilon}.
     \label{eq:zero_RS_est}
\end{equation}
This~$0^{\text{th}}$-order estimator can be used even when~$g$ is non-differentiable to obtain first-order information, which can then be exploited by any gradient-based optimization technique, as shown by applications in machine learning~\cite{NEURIPS2020_6bb56208}, computer vision~\cite{lelidec2021rendering,petersen2022gendr} and robotics for the optimal control of non-smooth dynamical systems~\cite{lelidec2022leveraging,suh2022bundled}.

\subsection{Direct application of randomized smoothing to collision detection}
Randomized smoothing can be applied to collision detection by considering the witness points as functions of the configuration vector, i.e.,~$(\bm x^*_1(\bm q),\bm x^*_2(\bm q))$, regardless of the method used to compute them.
We choose to use the combination of the two complementary and state-of-the-art algorithms, the Gilbert, Johnson and Keerthi algorithm (GJK)~\cite{GJK88} and Expanding Polytope Algorithm (EPA)~\cite{van_den_bergen_proximity_2001}, to handle collision detection including penetration.
Given two convex shapes~$\Aone$ and~$\Atwo$ and a relative pose~$T(\bm q)$, both of these algorithms allow to compute
a pair of witness points~$(\bm x^*_1(\bm q), \bm x^*_2(\bm q))$.

To compute a~$0^{\text{th}}$-order estimator of the gradients of~\eqref{eq:collision_detection_problem}, we perturb~$M$ times the configuration vector by sampling from a distribution~$\mu$:
\begin{equation}
     {\nabla^{(0)}_{\bm q} \, \bm x^*_{i,\epsilon}(\bm q)} =  \frac{1}{M} \sum_{j=0}^M -\bm x^*_i(\bm q + \epsilon \bm z^{(j)}) \frac{\nabla \log \mu (\bm z^{(j)})^\top}{\epsilon},
\end{equation}
for~$i=1,2$.
As a consequence, this~$0^{\text{th}}$-order estimator requires to run the GJK+EPA procedure~$M$ times.
While the choice of distribution may lead to different convolution effects~\cite{lelidec2021rendering,petersen2022gendr}, we found that the standard Gaussian distribution is well adapted to adequately sample over $SE(3)$.
In Fig.~\ref{fig:zero_order_sampling}, we give an intuition of the smoothing which results from the~$0^{\text{th}}$-order estimator.
To keep the visualization simple, we fix the pose of~$\Aone$ (in green) and perturb the pose of~$\Atwo$.
The cloud of resulting witness-points captures the local geometry of~$\Atwo$. Finally, note that finite differences is a sub-case of randomized smoothing, which considers a specific non-smooth distribution. By construction, finite differences capture less of the underlying geometry, as it deterministically defines the sampling directions and size of the steps taken in such directions. These fundamental differences are quantitatively illustrated in Table~\ref{fig:all_methods_cost_vs_iteration} of Sec.~\ref{sec:experiments}.
\section{$1^{\text{ST}}$-order estimation of collision detection derivatives}
\label{sec:first_order_diff}
\label{sec:implicit_differentiation}
In this section, we exploit the optimality conditions of the collision detection problem in order to introduce a computationally efficient and generic approach to derive a $1^\text{st}$-order estimate of the collision detection derivatives.
Similarly to the $0^\text{th}$-order approach developed in Sec.~\ref{sec:zero_order_diff}, we leverage randomized smoothing to compute the local Hessian information around the witness points required in the computation of specific gradient quantities.
In the particular case of meshes, we notably introduce a closed-form Hessian expression which can be directly computed from the vertices located in the neighborhood of the witness points.

\vspace{0.1cm}
\noindent
\textbf{The GJK paradigm for collision detection.}
To handle the problem of collision detection, we put ourselves in the paradigm of GJK~\cite{GJK88,montaut_collision_2022}, to which EPA~\cite{van_den_bergen_proximity_2001} also belongs.
In this paradigm, the focus is set on computing the~\textit{separation vector} between~$\Aone$ and~$\Atwo$.
The separation vector is defined as the smallest translation which must be applied to the relative configuration between~$\Aone$ and~$\Atwo$ in order to (i)~bring the shapes into collision if they are not in collision, or (ii)~separate the shapes if they are in collision.
Whether or not the shapes are in collision, the separation vector~\textit{always} satisfies the following equation:
\begin{equation}
    \begin{aligned}
      \xstar = \bm x^*_1 - \bm x^*_2,
    \end{aligned}
    \label{eq:separation_vector_and_witness_points}
\end{equation}
where~$\bm x^*_1$ and~$\bm x^*_2$ are the witness points, obtained as a by-product of the GJK and EPA algorithms.

Although this change in paradigm seems anecdotal, the problem of computing the separation vector~$\bm x^*$ between~$\Aone$ and~$\Atwo$ is conveniently fully encapsulated in a minimization problem over the Minkowski difference of the two shapes \mbox{$\Ddiff = \Aone - \Atwo = \{\bm x = \bm x_1 - \bm x_2 ~ | ~ \bm x_1 \in \Aone, \bm x_2 \in \Atwo\}$}:
\begin{equation}
    \begin{aligned}
      \xstar = & \argmin \norm{\bm x}^2_2 \\
              &\text{s.t. } \bm x \in \delta \Ddiff,
    \end{aligned}
    \label{eq:separation_vector_minimization_problem}
\end{equation}
where~$\delta \Ddiff$ is the boundary of the Minkowski difference.
Remarkably, the shapes are in collision if and only if the origin lies inside the Minkowski difference, i.e.,~$\bm 0 \in \Ddiff$, as shown in the seminal work of~\cite{GJK88} and revisited in~\cite{montaut_collision_2022}.
In summary, the separation vector is obtained by projecting the origin onto the surface of the Minkowski difference.

\vspace{0.1cm}
\noindent
\textbf{Optimality conditions of collision detection.}
In order to solve~\eqref{eq:separation_vector_minimization_problem}, GJK and EPA both solve a sequence of simple linear programming sub-problems.
Each sub-problem consists in computing the so-called~\textit{support function} of~$\Ddiff$:
\begin{equation}
  \suppDdiff(\bm x) = \max_{\bm y \in \Ddiff} \langle \bm y, \bm x \rangle,
  \label{eq:support_function_mink}
\end{equation}
where~$\bm x \in \Rthree$,~$\bm y \in \Ddiff$ and~$\langle \cdot, \cdot \rangle$ is the Euclidian dot-product.
A support point~$\bm s \in \delta \Ddiff$ belongs to the support set~$\partial \suppDdiff(\bm x)$, corresponding to the sub-gradient of~$\suppDdiff(\bm x)$, if and only if, it is a maximizer of~\eqref{eq:support_function_mink}.
Such a point always exists and is~\textit{always} a point on the boundary of the Minkowski difference.

Computing~$\suppDdiff(\bm x)$ corresponds to minimizing a linearization of the objective function of \eqref{eq:separation_vector_minimization_problem} at point~$-\bm x$.
Remarkably, when it is not possible to find a point~$\bm x \in \Ddiff$ which further decreases this linearization, both GJK and EPA have reached the optimal solution~$\bm x^*$, as Pb.~\eqref{eq:separation_vector_minimization_problem} is convex.
This optimality condition corresponds to the convergence criterion of both GJK~\cite{GJK88} and EPA~\cite{van_den_bergen_proximity_2001} and can be stated as follows:
\begin{equation}
  \begin{aligned}
    % & \langle \bm x^* - \bm s, \bm x^* \rangle = 0 \\
    % \iff & \langle \bm x^*, \bm x^* \rangle = \langle \bm x^*, \bm s \rangle = \suppDdiff(\bm x^*) \\
    \left\{
    \begin{array}{ll}
        \bm x^* \in \partial \suppDdiff(-\bm x^*) & \text{if } \bm 0 \not\in \Ddiff,\\
        \bm x^* \in \partial \suppDdiff(\bm x^*) & \text{otherwise}.
    \end{array}
    \right.
  \end{aligned}
  \label{eq:implicit_equation}
\end{equation}
Eq.~\eqref{eq:implicit_equation} is directly linked to the Frank-Wolf duality-gap, a convergence criteria which allows to measure the progress towards an optimal solution.
We refer to~\cite{montaut_collision_2022} for a complete analysis.
Eq.~\eqref{eq:implicit_equation} is handy as it corresponds to the optimality condition of~\eqref{eq:separation_vector_minimization_problem} and characterizes~$\bm x^*$.
Note that we choose to compute~$\bm x^*$ using GJK and EPA but~\eqref{eq:implicit_equation} is true however~$\bm x^*$ is obtained.
For the sake of simplicity, we will assume~\mbox{$\bm 0 \not\in \Ddiff$} but the rest of this section is applicable to the case where~\mbox{$\bm 0 \in \Ddiff$}.

For now, let us suppose first that~$\Ddiff$ is smooth and strictly-convex, which corresponds to the case where the shapes~$\Aone$ and~$\Atwo$ are smooth and strictly-convex (i.e., ellipsoids or spheres).
The support set is reduced to a singleton for any~$\bm x$:\mbox{~$\partial \suppDdiff(\bm x) = \nabla \suppDdiff(\bm x)$} where~$\nabla \suppDdiff$ is the gradient of the support function, i.e., the only maximizer of~\eqref{eq:support_function_mink}.
In such a case,~\eqref{eq:implicit_equation} reduces to an equality.

In practice, we use the support functions of the shapes denoted~$\suppAone$ and~$\suppAtwo$ to compute~$\suppDdiff$, as $\suppDdiff = \suppAone - \suppAtwo$~\cite{GJK88,montaut_collision_2022}.
Since we consider the relative pose~$T(\bm q)$ between~$\Aone$ and~$\Atwo$, all vectors are classically expressed in the frame of~$\Aone$.
By decomposing the terms in the support function, we can show that for any~$\bm x$, if~$\bm s_1 \in \partial \suppAone(\bm x)$ and~$\bm s_2 \in \partial \suppAone(-R(\bm q)^T \bm x)$, then:
\begin{equation}
    \bm s = \bm s_1 - \bm s_2 \in \partial \suppDdiff(\bm x),
    \label{eq:support_decomposition}
\end{equation}
where~$R(\bm q)$ is the rotation matrix associated to~$T(\bm q)$.

In practice, evaluating and finding a maximizer of the support function
is simple and a computationally cheap operation (this partly explains the large popularity of GJK and related algorithms for collision detection~\cite{ericsonRealTimeCollisionDetection}).
Since this is true also for~$\bm x^*$, we rewrite Eq.~\eqref{eq:implicit_equation} to obtain:
\begin{equation}
    \begin{aligned}
      \xstar - \nabla \suppAone(-\xstar) + T(\bm q)\nabla \suppAtwo(R(\bm q)^T \xstar) = \bm 0.
    \end{aligned}
    \label{eq:optimality_equation_shapes}
\end{equation}
Remarkably, the witness points~$\bm x^*_1$ and~$\bm x^*_2$ appear in~\eqref{eq:optimality_equation_shapes}:
\begin{equation}
    \begin{aligned}
    \left\{
        \begin{array}{ll}
          \bm x^*_1(\bm x^*) = \nabla \suppAone(-\bm x^*)& \\
          \bm x^*_2(\bm x^*, \bm q) = T(\bm q)\nabla \suppAtwo(R(\bm q)^\top \bm x^*).&
        \end{array}
    \right.
    \end{aligned}
    \label{eq:witness_points}
\end{equation}

\vspace{0.1cm}
\noindent
\textbf{Implicit function differentiation.}
We define the function~\mbox{$f: \Rthree \times \mathbb{R}^7 \rightarrow \Rthree$} as:
\begin{equation}
    f(\bm x, \bm q) = \bm x - \nabla \suppAone(-\bm x) + T(\bm q)\nabla \suppAtwo(R(\bm q)^T \bm x).
    \label{eq:implicit_function}
\end{equation}
From~\eqref{eq:optimality_equation_shapes}, the separation vector~$\bm x^*(\bm q)$, parameterized by~$\bm q$, is thus implicitly described by the equation:
\begin{equation}
    f(\bm x^*, \bm q) = \bm 0.
    \label{eq:implicit_function_equation}
\end{equation}
By expanding the~$1^{\text{st}}$-order terms of~$f$, we can relate the sensitivity of~$\bm x^*$ to the relative configuration~$\bm q$ between the shapes:
\begin{equation}
    \begin{aligned}
      \frac{\partial f(\xstar, \bm q)}{\partial \xstar} \delta \xstar + \frac{\partial f(\xstar, \bm q)}{\partial \bm q} \delta \bm q = \bm 0,
    \end{aligned}
    \label{eq:implicit_derivation}
\end{equation}
leading to the following relation:
\begin{equation}
    \begin{aligned}
      \frac{\partial f(\xstar, \bm q)}{\partial \xstar} \frac{\partial \xstar}{\partial \bm q} = -\frac{\partial f(\xstar, \bm q)}{\partial \bm q}, 
    \end{aligned}
    \label{eq:implicit_derivation_system}
\end{equation}
and finally to:
\begin{equation}
    \begin{aligned}
       \frac{\partial \xstar}{\partial \bm q} = 
       -\left[\frac{\partial f(\xstar, \bm q)}{\partial \xstar} \right]^{-1}
       \frac{\partial f(\xstar, \bm q)}{\partial \bm q}, 
    \end{aligned}
    \label{eq:implicit_derivation_system}
\end{equation}
if the Jacobian of $f$ w.r.t. $\xstar$ is invertible, where:
\begin{equation}
    \begin{aligned}
        \frac{\partial f(\xstar, \bm q)}{\partial \xstar} 
        = \mathit{I}  
        + \frac{\partial^2 \suppAone(-\bm x^*)}{\partial \bm x^{*2}} 
        + R(\bm q)\frac{\partial^2 \suppAtwo(\bm y^*)}{\partial \bm y^{*2}} R(\bm q)^{\top},
    \end{aligned}
    \label{eq:implicit_derivation_system_term_xstar}
\end{equation}
with~$\bm y^* = R(\bm q)\bm x^*$.
The terms in~$\partial f(\bm x^*, \bm q) / \partial \bm q$ are derivative terms involving elements of~$SE(3)$ and can be simply obtained by following the derivations in~\cite{sola_lie_2021}.

To evaluate both derivative terms of~$f$, it is necessary to compute the Hessian of the support function~$\partial^2 \suppA(\bm x) / \partial \bm x^2$ at~$\bm x^*$ (or~$\bm y^*$) where~$\mathcal{A}$ stands for either shape~$\Aone$ or~$\Atwo$.
% The derivative terms involving elements of~$SE(3)$ can be easily obtained by following the computations in~\cite{sola_lie_2021}.
The Hessian of the support function encodes the local curvature of the shape and it is easy to compute for basic smooth and strictly-convex shapes such as spheres or ellipsoids.

Let us now focus on the general case where~$\Aone$ or~$\Atwo$ might not be smooth or simply convex.
We explain how we can recover and compute the terms of~\eqref{eq:implicit_derivation_system}.
In the general case, Eq.~\eqref{eq:optimality_equation_shapes} becomes:
\begin{equation}
    \xstar - \bm x^*_1(\bm x^*) + T(\bm q)\bm x^*_2(\bm x^*, \bm q) = 0,
\end{equation}
with:
\begin{equation*}
    \begin{aligned}
        \left\{
        \begin{array}{ll}
          \bm x^*_1(\bm x^*) \in \partial \suppAone(-\bm x^*),& \\
          \bm x^*_2(\bm x^*, \bm q) \in \partial \suppAtwo(R(\bm q)^\top \bm x^*),&
        \end{array}
        \right.
    \end{aligned}
\end{equation*}
which are computed by the GJK+EPA procedure. By casually writing~\mbox{$\nabla \suppAone(-\bm x^*) = \bm x^*_1(\bm x^*)$} and~\mbox{$\nabla \suppAtwo(R(\bm q)^T \bm x^*_2) = \bm x^*_2(\bm x^*, \bm q)$}, we recover~\eqref{eq:optimality_equation_shapes} and ultimately~\eqref{eq:implicit_derivation_system}.

\vspace{0.1cm}
\noindent
\textbf{Uninformative gradients: the case of meshes.}
When a shape is non-smooth or non-strictly convex, the Hessian of its support function may be null (e.g., a flat surface) or even undefined (e.g., the Hessian at the vertex of a cube).
Let us consider the example of a mesh representing the convex-hull of an arbitrary object to illustrate this phenomenon.
A mesh~$\mathcal{A}$ is a list of~$N_v$ vertices~$\{\bm v_1, ... \bm v_{N_v}\}$ and therefore:
\begin{equation}
    \begin{aligned}
        &\suppA(\bm x) = \max_{\bm v_i \in \{\bm v_1,...,\bm v_{N_v}\}} \langle \bm v_i, \bm x \rangle,\\
        &\nabla \suppA(\bm x) = \bm v_{i^*}, 
    \end{aligned}
    \label{eq:supp_mesh}
\end{equation}
for a certain~$i^* \in [1, N_v]$.
We define the matrix~\mbox{$V \in \mathbb{R}^{3\times N_v}$} and the vector~\mbox{$\bm z(\bm x) \in \mathbb{R}^{N_v}$} as:
\begin{equation}
    \begin{aligned}
        & V^T = (\bm v_1 \cdot \cdot \cdot \bm v_{N_v})^T,\\
        & \bm z(\bm x) = V^T \bm x.
    \end{aligned}
    \label{eq:vertex_matrix}
\end{equation}
This allows us to define the vector of~\textit{argmax weights}~\mbox{$\bm a(\bm z) \in \mathbb{R}^{N_v}$}:
\begin{equation}
     \bm a(\bm z) = \argmax_{\norm{\bm w}_1 \leq 1,~\bm 0 \leq \bm w} \bm z^T \bm w.
    \label{eq:argmax_weights}
\end{equation}
Therefore, all components of~$\bm a(\bm z(\bm x))$ are null, except the ${i^*}^\text{th}$ component which value is~$1$.
As a consequence, we have:
\begin{equation}
    \nabla \suppA(\bm x) = \bm v_{i^*} = \sum_{i} a_i \bm v_i = V \bm a(\bm z(\bm x)),
    \label{eq:supp_and_argmax_weights}
\end{equation}
and:
\begin{equation}
    \frac{\partial^2 \suppA(\bm x)}{\partial \bm x^2} =
    V \frac{\partial \bm a(\bm z)}{\partial \bm z} V^{\top} = \bm 0,
    \label{eq:hessian_supp_mesh}
\end{equation}
because~$\partial \bm a(\bm y) / \partial \bm y$ is null almost everywhere.
As a consequence, the gradients of~$\bm x^*$ w.r.t~$\bm q$ obtained after solving~\eqref{eq:implicit_derivation_system} fail to capture information regarding the curvature of the underlying object which the mesh approximates.

%%%%%%%%%% FIGURE %%%%%%%%%%
%%%%%%%%%%%%%%%%%%%%%%%%%%%%
\begin{figure}[!t]
    \vspace{-0.5cm}
    \centering
    \subfloat[
        Finite differences.
        \label{fig:}
    ]{
        \includegraphics[width=0.45\linewidth]{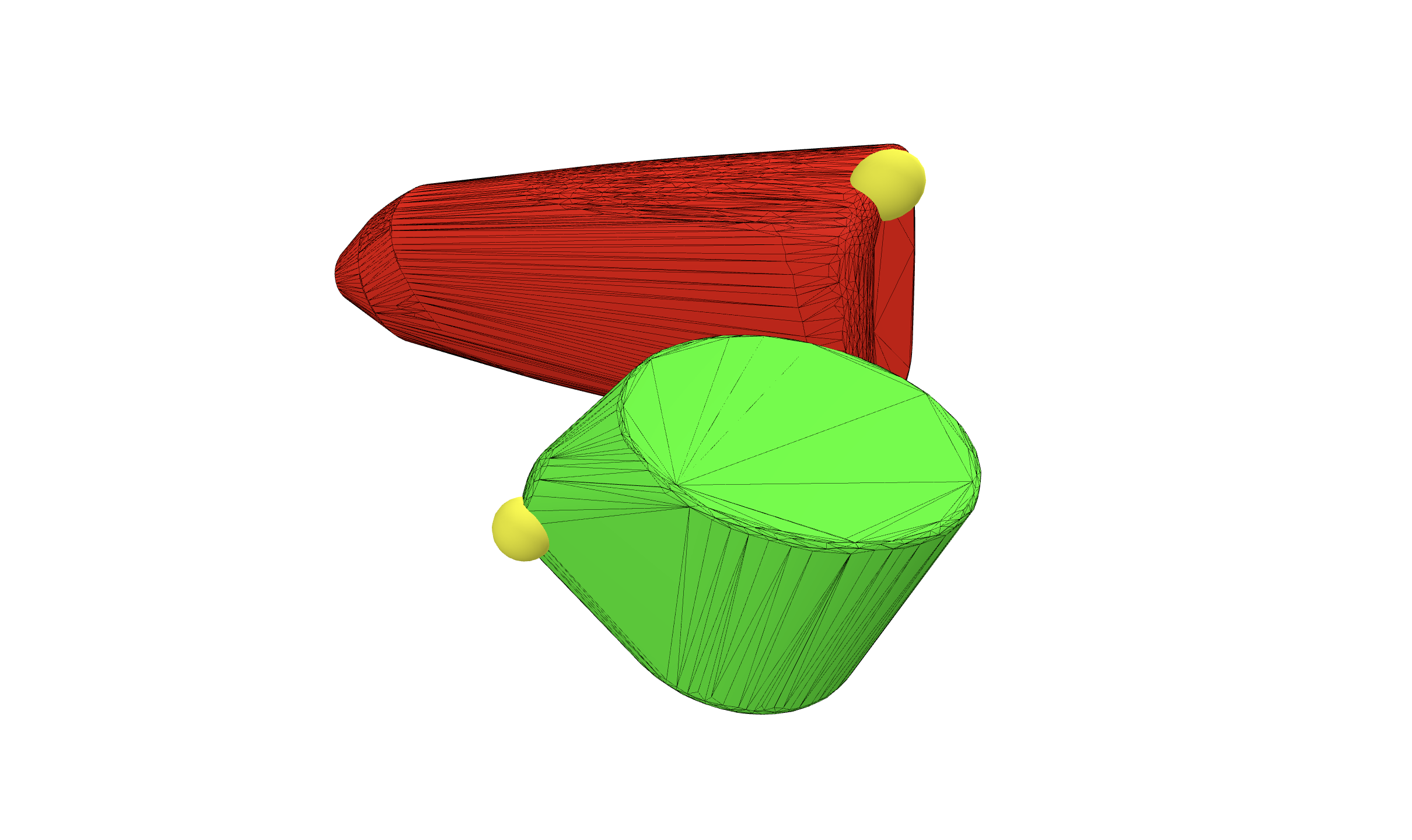}
    }
    % \hspace{1cm}
    \subfloat[
        $0^{\text{th}}$-order.
        \label{fig:}
    ]{
        \includegraphics[width=0.45\linewidth]{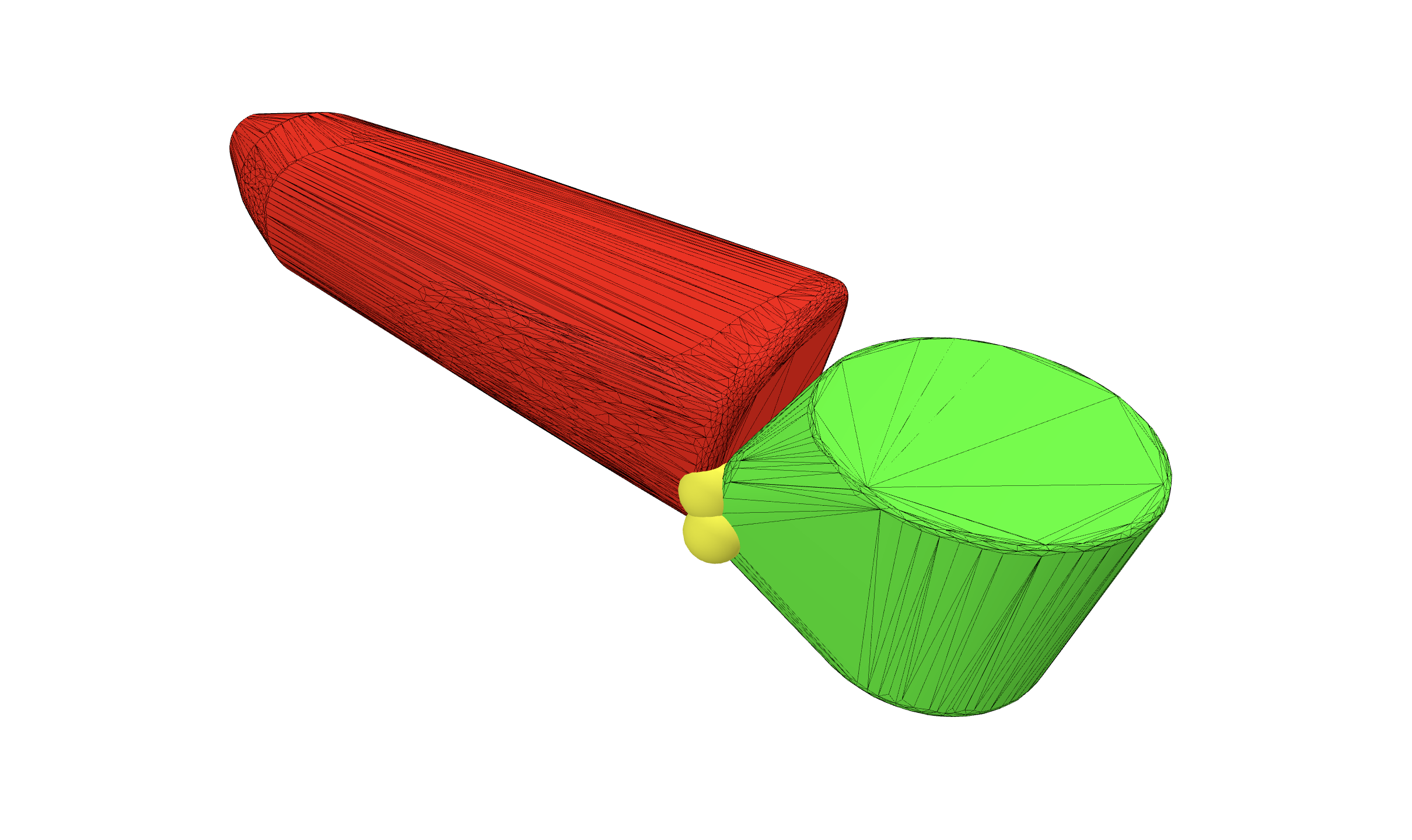}
    }\\
    % \hspace{1cm}
    \subfloat[
        $1^{\text{st}}$-order. 
        Gumbel and Gaussian.
        \label{fig:}
    ]{
        \includegraphics[width=0.3\linewidth]{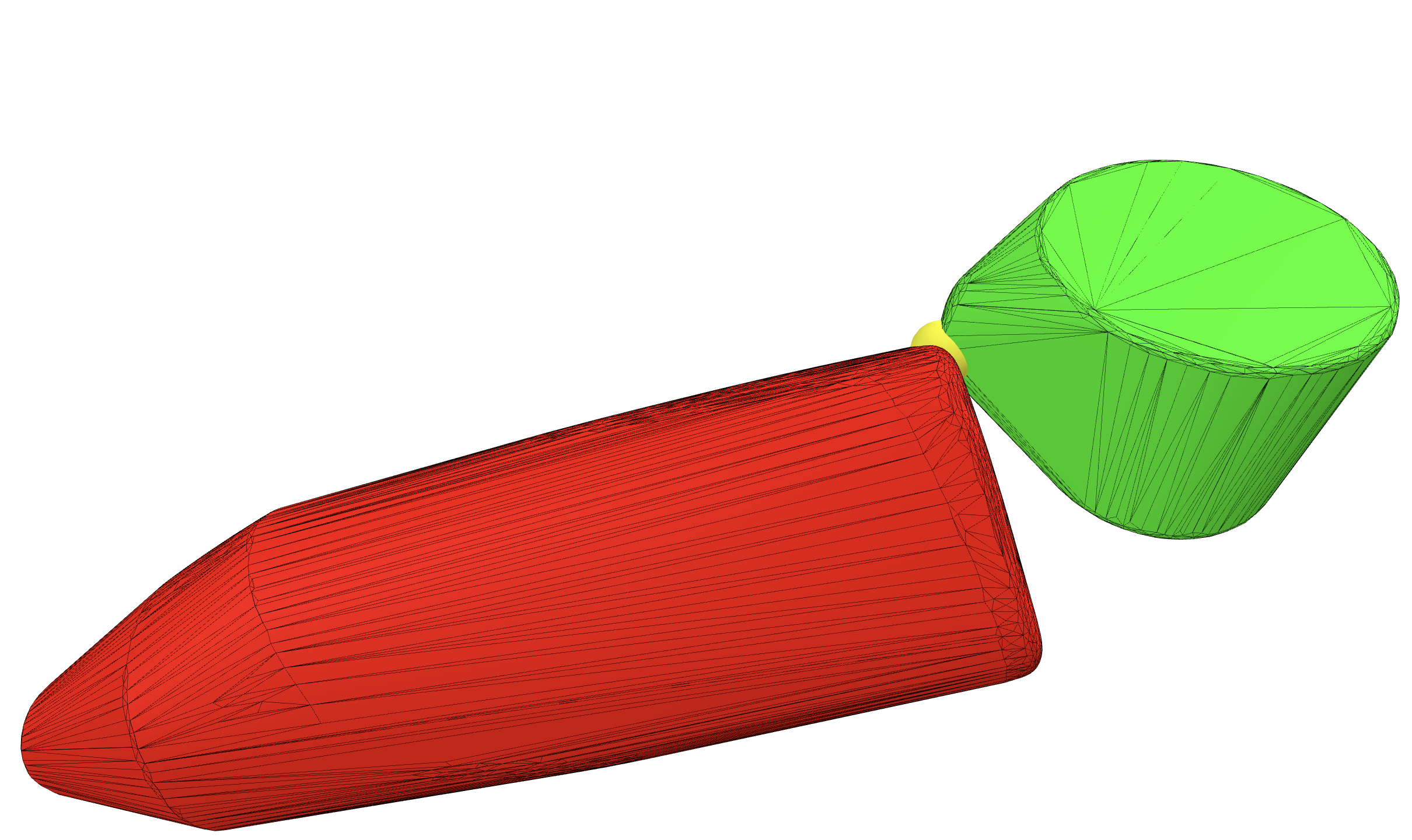}
    }
    \caption{\small
        \textbf{Convex hulls of YCB shapes, goal: find a relative pose such that the yellow points are contact points.} 
        We display the final pose each method converged to.
        Finite differences fail to generate a satisfying pose.
        The~$0^{th}$-order gradient estimator is better than finite differences but is not as precise as the~$1^{st}$-order gradient estimator.
    }
    \label{fig:all_methods_final_pose}
    \vspace{-0.5cm}
\end{figure}
%%%%%%%%%%%%%%%%%%%%%%%%%%%%
%%%%%%%%%%%%%%%%%%%%%%%%%%%%

\vspace{0.1cm}
\noindent
\textbf{Hessian estimation via randomized smoothing.}
To overcome this issue, we use a randomized smoothing approach in order to estimate~$\partial^2 \suppA(\bm x) / \partial \bm x^2$.
We apply~\eqref{eq:zero_RS_est} to~$g = \nabla \suppA$ to obtain:
\begin{equation}
    \frac{\partial^2 \suppA (\bm x)}{\partial \bm x^2} \approx 
     \frac{1}{M} \sum_{j=0}^M -\nabla \suppA(\bm x + \epsilon \bm z^{(j)}) \frac{\nabla \log \mu (\bm z^{(j)})^\top}{\epsilon}.
     \label{eq:first_RS_est_hessian_support}
\end{equation}
In practice, we choose~$\mu$ to be a normalized Gaussian centered in~$\bm 0$.
Although this procedure to evaluate the Hessian of the support function requires to compute the support function~$M$ times, it is in practice very efficient, as first the computation of the support function is very cheap, second it can be warm-started and third it is highly parallelisable.
Finally, this method to estimate the Hessian of a support function is generic as it can be applied to \textit{any} convex shape with a computationally tractable support function.

\vspace{0.1cm}
\noindent
\textbf{Special case of meshes: the Gumbel distribution.}
In the specific case of meshes, the structure of the support function allows to replace the Gaussian distribution by the Gumbel distribution~\cite{gumbel1954statistical}, resulting in a closed form solution to estimate the mean of~$\partial^2 \suppA(\bm x) / \partial \bm x^2$ and thus removing the need of a Monte-Carlo estimator.
Thus, using a Gumbel distribution~$\mu$ with zero mean and identity matrix variance to sample the noise, we get a closed-form solution for~\eqref{eq:rs_smooth_function} when applied to~$g(\bm z) = \bm a(\bm z)$ from Eq.~\eqref{eq:argmax_weights}:
\begin{equation}
    \begin{aligned}
        \bm a_{\epsilon}(\bm z) &= \mathbb{E}_{Z\sim \mu} \left[ a(\bm z + \epsilon Z) \right] \\
        &= 
        \frac{1}{\sum_{j} e^{z_j/\epsilon}}
        \left(
        e^{z_1/\epsilon}
        ~\ldots
        ~e^{z_{N_v}/\epsilon}
        \right)^{\top}.
    \end{aligned}
    \label{eq:soft_argmax}
\end{equation}
The smoothed~$\bm a_{\epsilon}$ is thus simply a soft-max, which is a smooth and differentiable function of~$\bm y$~\cite{NEURIPS2020_6bb56208}.
The~$\epsilon$ parameter serves as a temperature parameter~\cite{gumbel1954statistical}.
Due to the nature of the soft-max operator, the~$i^{th}$ weight in~$\bm a_{\epsilon}$ decreases exponentially the further~$z_i = \langle \bm v_i, -\bm x^*\rangle$ is from the maximum value~$\suppA(-\bm x^*)$.
This maximum value is attained by the witness point of the considered shape~$\mathcal{A}$.
We illustrate in Fig.~\ref{fig:first_order_gumbel_neighbors} the weighting of this soft-max operation on a mesh.
Remarkably, the further a vertex is from the current witness point, the less it contributes to the soft-max.
It is therefore only necessary to keep the points of the mesh which belong to a neighborhood around the witness point of shape~$\mathcal{A}$.
This fact renders the use of the Gumbel distribution very efficient on meshes.
By choosing the~\textit{depth} of neighboring vertices around the witness point, which we denote by~$n_l$, we can choose to limit or increase the number of neighbors involved in the computation of~\eqref{eq:soft_argmax}.
As an example, a depth of~$n_l=2$ corresponds to keeping only the neighbors of the witness points and the neighbors of neighbors.

Finally, by applying the chain rule we get the estimation of~$\partial^2 \suppA(\bm x) / \partial \bm x^2$:
\begin{equation}
    \frac{\partial^2 \suppA(\bm x)}{\partial \bm x^2}
    \approx
    \frac{\partial \bm a_{\epsilon}(\bm z(\bm x))}{\partial \bm x}
    =
    V \frac{\partial \bm a_{\epsilon}(\bm z)}{\partial \bm z} V^T,
    \label{eq:hessian_support_mesh}
\end{equation}
where~$\partial \bm a_{\epsilon}(\bm z)/\partial \bm z$ is simply the derivative of the soft-max function.

\vspace{0.1cm}
\noindent
\textbf{Derivatives of the witness points.}
In general,~\mbox{$\bm x^*_1 = \nabla \suppAone(-\bm x^*)$} so by applying the chain rule we get:
\begin{equation}
    \frac{\partial \bm x^*_{1,2}}{\partial \bm q}
    =
    - \frac{\partial \nabla \sigma_{\mathcal{A}_{1,2}}(-\bm x^*)}{\partial \bm x^*}
    \frac{\partial \bm x^*}{\partial \bm q}
    = 
    - \frac{\partial^2 \sigma_{\mathcal{A}_{1,2}}(-\bm x^*)}{\partial \bm x^{*2}}
    \frac{\partial \bm x^*}{\partial \bm q}
    .
    \label{eq:derivative_support}
\end{equation}

To conclude this section, we have introduced a complete approach to retrieve a $1^\text{st}$-order estimate of the variation of the witness points lying on the two shapes, according to the relative placements between these two.
In particular, in the case of meshes, we have proposed a closed-form formula to compute a local approximation of the Hessian using the neighborhood of the current witness points.
\section{Experiments}
\label{sec:experiments}
In this section, we evaluate whether the gradients obtained with the two proposed estimators are meaningful and how computationally efficient it is to compute them compared to finite differences.

\vspace{0.1cm}
\noindent
\textbf{Contact-pose generation benchmark.}
To answer the first question, we evaluate the~$0^{\text{th}}$-order and~$1^{\text{st}}$-order estimators against finite differences on a synthetic benchmark of non-trivial minimization problems.
%%%%%%%%%% FIGURE %%%%%%%%%%
%%%%%%%%%%%%%%%%%%%%%%%%%%%%
\begin{figure}[!t]
    \vspace{-0.5cm}
    \centering
    \includegraphics[width=0.5\linewidth]{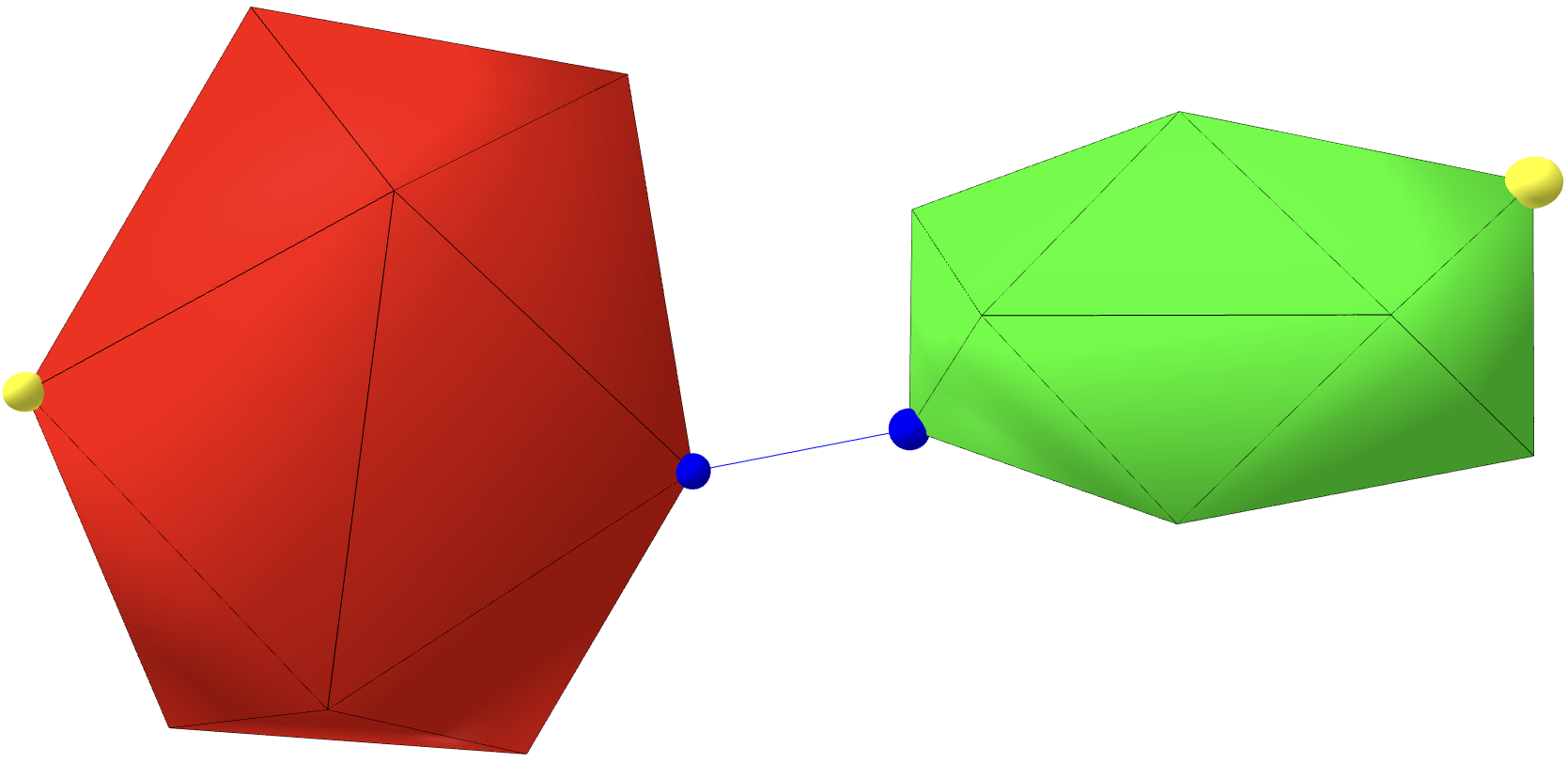}
    \caption{\small\textbf{Rough shapes, goal: find a relative pose such that the yellow points are contact points.}
    The current witness-points are the blue points.
    }
    \label{fig:poly_ellipsoids}
    \vspace{-0.5cm}
\end{figure}
%%%%%%%%%%%%%%%%%%%%%%%%%%%%
%%%%%%%%%%%%%%%%%%%%%%%%%%%%
We generate~$100$ collision pairs with random polyhedral ellipsoids, i.e., ellipsoids which surfaces are represented by a~$12$-vertices convex mesh (see Fig.~\ref{fig:poly_ellipsoids}).
The resulting shapes are rough, i.e., the curvature information of the original shape has been greatly truncated.
For each pair of convex shapes~$(\Aone, \Atwo)$, we generate~$100$ random target points~$\bm x_{1, \text{des}} \in \Aone$ and~$\bm x_{2, \text{des}} \in \Atwo$ on the shapes' surfaces.
For each of the~$10000$ generated problems, the goal is to find a relative pose~$T(\bm q)$ between~$\Aone$ and~$\Atwo$ such that the shapes are in contact and their witness points~$\bm x^*_1(\bm q)$ and~$\bm x^*_2(\bm q)$ satisfy~$\bm x^*_1(\bm q) = \bm x_{1, \text{des}}$ and~$\bm x^*_2(\bm q) = \bm x_{2, \text{des}}$.
Mathematically, this corresponds to solving the minimization problem:
\begin{equation}
    \begin{aligned}
    \min_{\bm q} 
        \frac{1}{2}\sum_{i=1,2} \norm{\bm x^*_i(\bm q) - \bm x^*_{i, \text{des}}}^2
        + \frac{1}{2} \norm{\bm x^*_1(\bm q) - \bm x^*_2(\bm q)}^2.
    \end{aligned}
    \label{eq:experiments_minimization_problem}
\end{equation}
To solve this minimization problem, we use the Gauss-Newton algorithm with backtracking line search~\cite{wright1999numerical} and run it for~$50$ iterations.
To compute the Jacobian of the cost, we evaluate the terms~$\partial \bm x^*_{1,2} / \partial \bm q$, with finite differences, and the~$0^{\text{th}}$ and~$1^{\text{st}}$-order estimators proposed in this work.
Finally, to compute~$\bm x^*_1(\bm q)$ and~$\bm x^*_2(\bm q)$, we use the combination of the GJK and EPA algorithms implemented in the HPP-FCL library~\cite{mirabel2016hpp,hppfclweb}, a fork of the FCL library~\cite{panFCLGeneralPurpose2012}.

We report as quantiles in Table~\ref{tab:poly_ellipsoid_benchmark} the value of the terminal cost~$C(\bm q)$.
The lower this quantity, the better the quality of the solution found.
We are particularly interested in the quantiles Q3 and D9 of Table~\ref{tab:poly_ellipsoid_benchmark}: respectively~$25\%$ and~$10\%$ of problems have a terminal cost worse (higher) than the reported value.
The higher this value is, the less reliable the method is.
We observe that finite differences have a high value for Q3 and D9 whereas the two proposed estimators are at least one order of magnitude better.
Remarkably, the~$1^{\text{st}}$-order estimators, whether the underlying distribution used is Gaussian or Gumbel, is extremely accurate and reliable, with a value of Q3 and D9 at least 3 orders of magnitude better than finite differences.

\begin{figure}
    % \vspace{-0.5cm}
    \centering
    \includegraphics[width=0.8\linewidth]{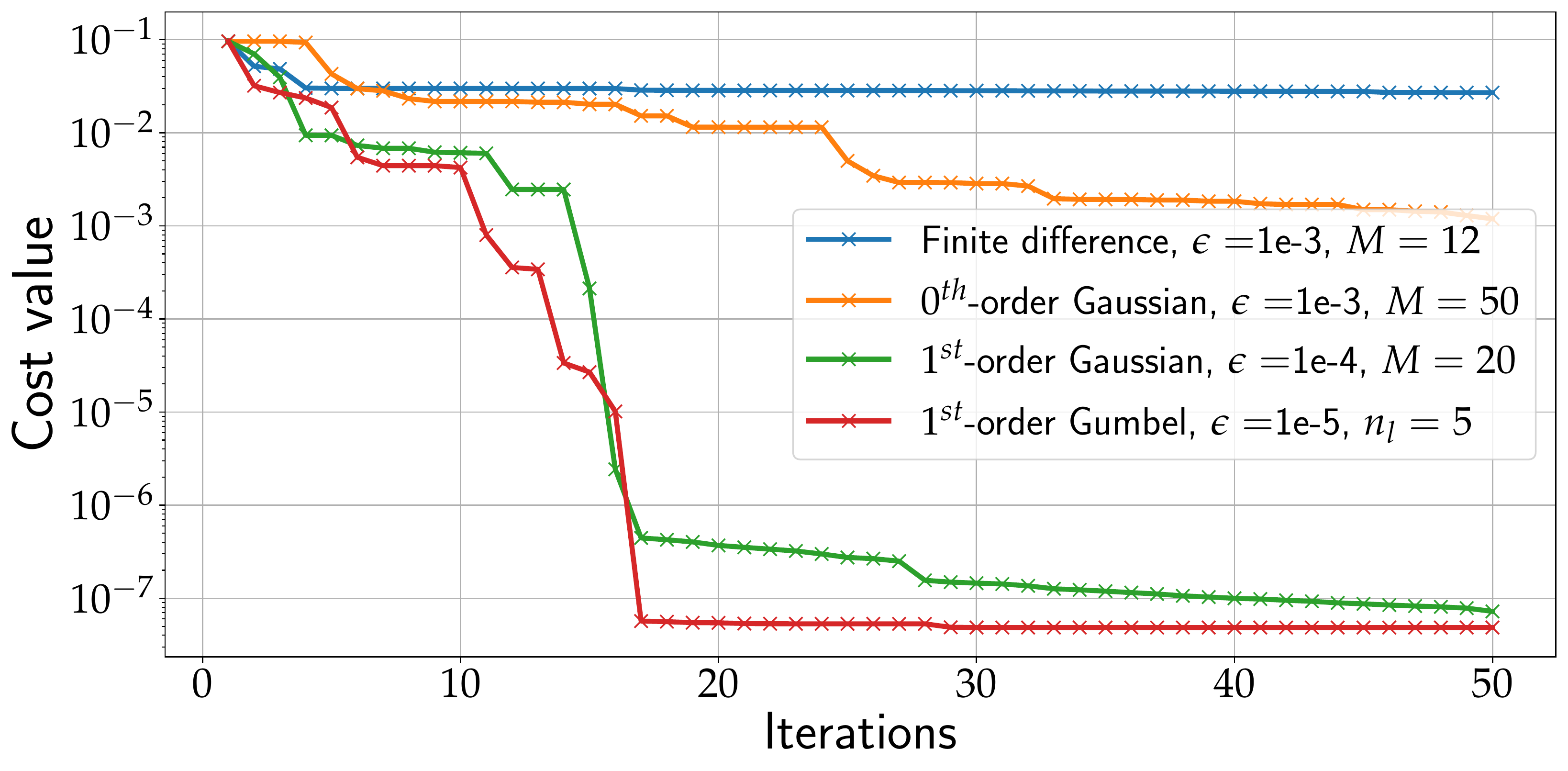}
    \caption{\small
        \textbf{YCB shape, goal: find a relative pose such that the yellow points are contact points.} 
        Finite difference typically gets stuck around a limited precision (depending on the finite difference increment).
        The~$1^{\text{st}}$-order estimator converges rapidly towards a solution with high precision unlike the~$0^{\text{th}}$-order estimator.
    }
    \label{fig:all_methods_cost_vs_iteration}
    \vspace{-0.5cm}
\end{figure}

\begin{figure}
    \centering
    \includegraphics[width=0.8\linewidth]{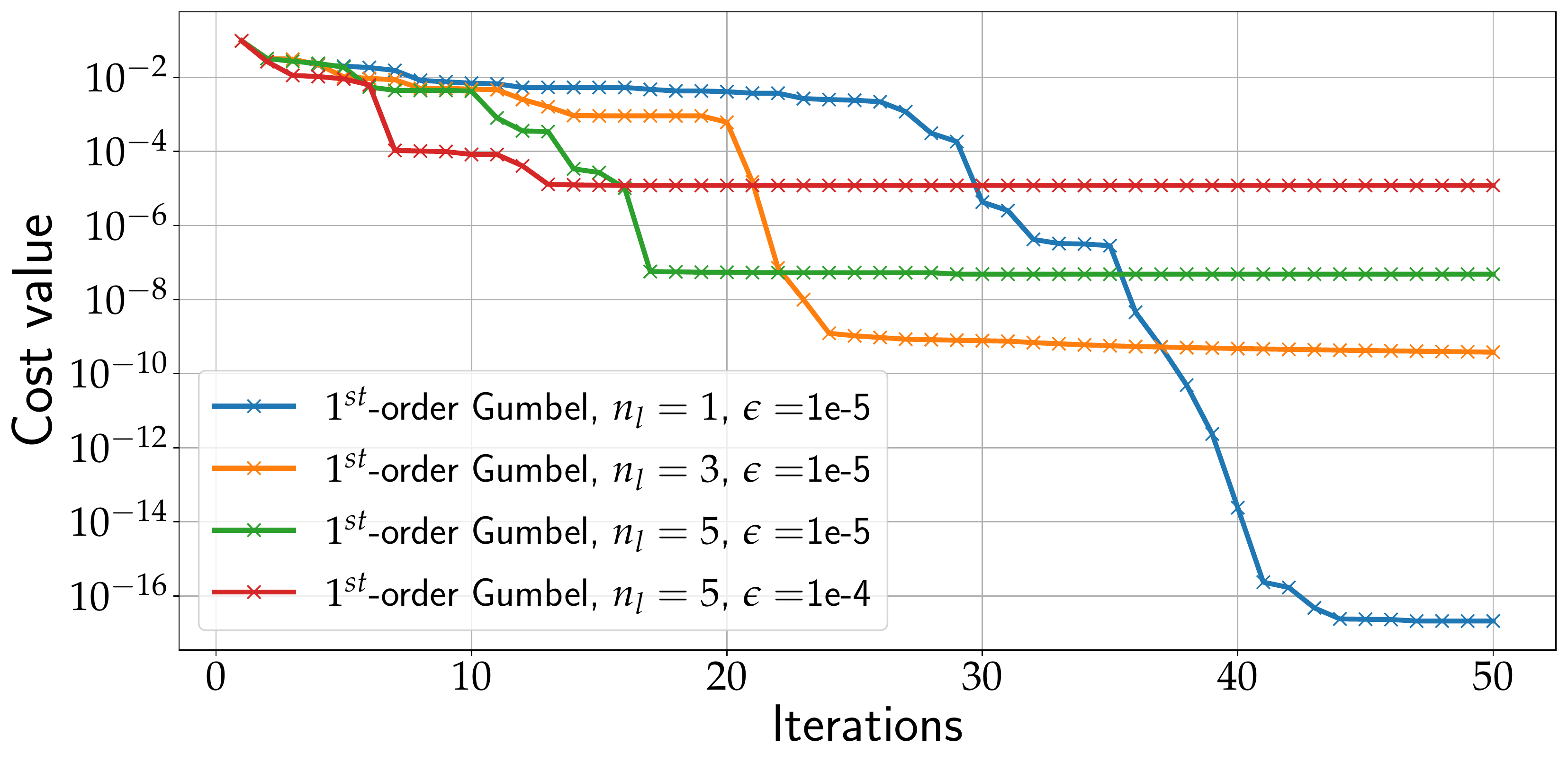}
    \caption{\small
        \textbf{Goal: find a relative pose such that the yellow points are contact points.} 
        With more noise, the faster the convergence is at first, but the estimator ends up being less precise, hence converges to a less optimal solution.
    }
    \label{fig:first_order_methods_cost_vs_iteration}
    \vspace{-0.3cm}
\end{figure}

As an additional qualitative example, Fig.~\ref{fig:all_methods_final_pose} and Fig.~\ref{fig:all_methods_cost_vs_iteration} show a typical example of failure of finite differences on a collision pair of the YCB dataset - a dataset which contains high-resolution meshes of real-world household objects~\cite{calli2015ycb}.
Finally, Fig.~\ref{fig:first_order_methods_cost_vs_iteration} shows the typical impact of the noise and number of samples on the~$1^{\text{st}}$-order estimator using the Gumbel distribution for the same YCB problem.
The same kind of behavior is obtained when using a Gaussian distribution.
Overall, the higher the number of samples, the better the quality of the estimator; the higher the noise, the faster the convergence at the price of reduced accuracy.
%%%%%%%%%% TABLE %%%%%%%%%%%
%%%%%%%%%%%%%%%%%%%%%%%%%%%%
\begin{table}[]
\vspace{-0.1cm}
\centering
\resizebox{\columnwidth}{!}{%
\begin{tabular}{c|cc|c|c|c}
       & \multicolumn{2}{c|}{\begin{tabular}[c]{@{}c@{}}Finite\\ differences\end{tabular}} & \begin{tabular}[c]{@{}c@{}}$0^{\text{th}}$-order\\ Gaussian\end{tabular} & \begin{tabular}[c]{@{}c@{}}$1^{\text{st}}$-order\\ Gaussian\end{tabular} & \begin{tabular}[c]{@{}c@{}}$1^{\text{st}}$-order \\ Gumbel\end{tabular} \\ \cline{2-6} 
       & \multicolumn{1}{c|}{\begin{tabular}[c]{@{}c@{}}$M=12$\\ $\epsilon=10^{-6}$\end{tabular}} & \multicolumn{1}{c|}{\begin{tabular}[c]{@{}c@{}}$M=12$\\ $\epsilon=10^{-3}$\end{tabular}} & \begin{tabular}[c]{@{}c@{}}$M=50$\\ $\epsilon=10^{-2}$\end{tabular} & \begin{tabular}[c]{@{}c@{}}$M=20$\\ $\epsilon=10^{-3}$\end{tabular}              & \begin{tabular}[c]{@{}c@{}}$n_l=1$\\ $\epsilon=10^{-4}$\end{tabular}             \\ \hline
D1     & $2\times 10^{-33}$ & $8\times 10^{-33}$  & $4\times 10^{-23}$ & $4\times 10^{-16}$ & $6\times 10^{-16}$ \\
Q1     & $4\times 10^{-32}$ & $7\times 10^{-23}$  & $5\times 10^{-20}$ & $1\times 10^{-10}$ & $3\times 10^{-10}$ \\
Median & $3\times 10^{-3}$  & $4\times 10^{-14}$  & $2\times 10^{-13}$ & $4\times 10^{-8}$  & $1\times 10^{-8}$ \\
Q3     & $4\times 10^{-2}$  & $1\times 10^{-2}$   & $2\times 10^{-3}$  & $3\times 10^{-7}$  & $7\times 10^{-8}$ \\
D9     & $8\times 10^{-2}$  & $5\times 10^{-2}$   & $5\times 10^{-3}$  & $2\times 10^{-6}$  & $2\times 10^{-5}$  
\end{tabular}%
}
\caption{\small
    \textbf{Rough shapes (see Fig.~\ref{fig:poly_ellipsoids}),
    value of~$C(\bm q)$ after 50 iterations of Gauss-Newton with line search.} 
    Q3 and D9: respectively~$25\%$ and~$10\%$ of problems have a terminal cost worse (higher) than the reported value.
}
\label{tab:poly_ellipsoid_benchmark}
\vspace{-0.1cm}
\end{table}
%%%%%%%%%%%%%%%%%%%%%%%%%%%%
%%%%%%%%%%%%%%%%%%%%%%%%%%%%

%%%%%%%%%% FIGURE %%%%%%%%%%
%%%%%%%%%%%%%%%%%%%%%%%%%%%%
%%%%%%%%%%%%%%%%%%%%%%%%%%%%
%%%%%%%%%%%%%%%%%%%%%%%%%%%%

%%%%%%%%%% FIGURE %%%%%%%%%%
%%%%%%%%%%%%%%%%%%%%%%%%%%%%
%%%%%%%%%%%%%%%%%%%%%%%%%%%%
%%%%%%%%%%%%%%%%%%%%%%%%%%%%

% \subsection{Timings benchmark}
\vspace{0.1cm}
\noindent
\textbf{YCB timings benchmarks.}
To evaluate the computational efficiency of the proposed estimators, we generate~$10000$ collision detection problems~\eqref{eq:collision_detection_problem} using meshes from the YCB real-world objects dataset~\cite{calli2015ycb} and measure the time taken to compute the derivatives of witness points for each estimator.
For the finite differences and~$0^{\text{th}}$-order, GJK+EPA are warm started at each sample~$\bm q + \epsilon \bm z$ to enhance the computational efficiency.
The results, reported in Table~\ref{tab:timings_table} show that although the~$0^{\text{th}}$-order estimator is often prohibitive compared to finite differences, the~$1^{\text{st}}$-order estimators using Gaussian and Gumbel distributions can be obtained extremely efficiently, on the order of the micro-seconds and from~$10$ to~$70$ times faster than finite differences.
%%%%%%%%%% TABLE %%%%%%%%%%%
%%%%%%%%%%%%%%%%%%%%%%%%%%%%
\begin{table}[!t]
\centering
% \resizebox{\columnwidth}{!}{%
\begin{tabular}{c|c|cc}
\multirow{2}{*}{Method} & \multirow{2}{*}{Parameters} & \multicolumn{2}{c}{Timings (in $\mu$s)}    \\ \cline{3-4} 
                        &                             & \multicolumn{1}{c|}{Collision} & No collision \\ \hline
\begin{tabular}[c]{@{}c@{}}Finite \\ differences\end{tabular}                 & $M=12$                     & $78\pm48$               & $11.0\pm7.3$     \\ \hline
\multirow{4}{*}{\begin{tabular}[c]{@{}c@{}}$0^{\text{th}}$-order\\ Gaussian\end{tabular}} & $M=10$           & $65\pm36$               & $9\pm6$          \\
                                                                              & $M=20$                & $122\pm130$             & $17\pm10$          \\
                                                                              & $M=50$                & $338\pm391$             & $47\pm30$         \\
                                                                              & $M=100$               & $613\pm402$             & $86\pm45$         \\ \hline
\multirow{4}{*}{\begin{tabular}[c]{@{}c@{}}$1^{\text{st}}$-order\\ Gaussian\end{tabular}} & $M=10$           & $4.8\pm1.4$             & $3.1\pm0.3$       \\
                                                                              & $M=20$                & $8.9\pm2.8$             & $5.8\pm0.5$       \\
                                                                              & $M=50$                & $22\pm8$                & $15\pm3$       \\
                                                                              & $M=100$               & $42\pm13$                & $27\pm3$          \\ \hline
\multirow{3}{*}{\begin{tabular}[c]{@{}c@{}}$1^{\text{st}}$-order\\ Gumbel\end{tabular}}& $n_l=1$             & $1.7\pm0.5$             & $1.6\pm0.5$        \\
                                                                              & $n_l=3$               & $4.1\pm1.8$             & $3.9\pm1.4$        \\
                                                                              & $n_l=5$               & $9.6\pm7.8$             & $9.7\pm8.1$         
\end{tabular}%
% } 
\caption{\small
    \textbf{Timings for computing collision detection derivatives, for collision pairs of the YCB dataset.} 
    The parameters were selected in the ranges which would typically be used in practice.
}
\label{tab:timings_table}
\vspace{-0.5cm}
\end{table}
%%%%%%%%%%%%%%%%%%%%%%%%%%%%
%%%%%%%%%%%%%%%%%%%%%%%%%%%%
\section{Conclusion}
\label{sec:conclusion}
In this paper, we propose a generic approach for computing $0^\text{th}$ and $1^\text{th}-$order derivatives of collision detection for \textit{any} convex shapes by leveraging randomized smoothing techniques. 
While being robust and easy to implement, our approach exhibits strong benefits in terms of speed and accuracy, taking only few micro-seconds to compute informative derivatives of complex shapes, such as meshes with hundred vertices, involved in real robotic applications. 
Remarkably, the $1^\text{th}-$order estimators are also both more accurate and less expensive to compute than the~$0^{\text{th}}$-order estimator.
All these gradient estimation methods have been implemented in the HPP-FCL and Pinocchio ecosystems.
We plan to extend our contributions by applying these methods for differentiable simulation, optimal grasp synthesis and trajectory optimization.

\clearpage
\balance 
\bibliographystyle{ieeetr}
\bibliography{references}

\end{document}